\documentclass[letterpaper]{article} 
\usepackage{aaai23}  
\usepackage{times}  
\usepackage{helvet}  
\usepackage{courier}  
\usepackage[hyphens]{url}  
\usepackage{graphicx} 
\usepackage{natbib}  
\usepackage{caption} 
\usepackage{algorithm}
\usepackage{algorithmic}
 
\usepackage{newfloat}
\usepackage{listings}
\DeclareCaptionStyle{ruled}{labelfont=normalfont,labelsep=colon,strut=off} 
\floatstyle{ruled}
\newfloat{listing}{tb}{lst}{}
\floatname{listing}{Listing}
\usepackage{bm}
\usepackage{amsmath}
\usepackage{amssymb}
\usepackage{multirow}
\usepackage{booktabs}
\usepackage{amsthm}
\usepackage{subfig}

\title{Rephrasing the Reference for Non-Autoregressive Machine Translation}

\author{Chenze Shao\textsuperscript{\rm{1,2}}, Jinchao Zhang\textsuperscript{\rm{3}}, Jie Zhou\textsuperscript{\rm{3}}, Yang Feng\textsuperscript{\rm{1,2}}\thanks{Joint work with Pattern Recognition Center, WeChat AI, Tencent Inc. Yang Feng is the corresponding author. Reproducible code: https://github.com/ictnlp/Rephraser-NAT.}}

\affiliations{ 
\textsuperscript{\rm{1}} Key Laboratory of Intelligent Information Processing,\\ Institute of Computing Technology, Chinese Academy of Sciences\\
\textsuperscript{\rm{2}} University of Chinese Academy of Sciences\\
\textsuperscript{\rm{3}} Pattern Recognition Center, WeChat AI, Tencent Inc, China\\
{ \{shaochenze18z,fengyang\}@ict.ac.cn}, 
{ \{dayerzhang,withtomzhou\}@tencent.com}
}

\begin{document}
\maketitle
\begin{abstract}
Non-autoregressive neural machine translation (NAT) models suffer from the multi-modality problem that there may exist multiple possible translations of a source sentence, so the reference sentence may be inappropriate for the training when the NAT output is closer to other translations. In response to this problem, we introduce a rephraser to provide a better training target for NAT by rephrasing the reference sentence according to the NAT output. As we train NAT based on the rephraser output rather than the reference sentence, the rephraser output should fit well with the NAT output and not deviate too far from the reference, which can be quantified as reward functions and optimized by reinforcement learning. Experiments on major WMT benchmarks and NAT baselines show that our approach consistently improves the translation quality of NAT. Specifically, our best variant achieves comparable performance to the autoregressive Transformer, while being 14.7 times more efficient in inference.
\end{abstract}

\section{Introduction}
Non-autoregressive neural machine translation \citep[NAT,][]{gu2017non} models significantly speed up the decoding but suffer from performance degradation compared to autoregressive models. The performance degradation is mainly attributed to the multi-modality problem that there may exist multiple possible translations of a source sentence, so the reference sentence will be inappropriate for the training when the NAT output is closer to other translations. The multi-modality problem manifests itself as the large cross-entropy loss during the training, which cannot evaluate the NAT output properly. To minimize the training loss, NAT tends to generate a mixture of multiple translations rather than a consistent translation, which typically contains many repetitive tokens in the generated results.

A number of efforts have explored ways to help NAT handle the multi-modality problem. A basic solution is sequence-level knowledge distillation \cite{kim-rush-2016-sequence,Zhou2020Understanding}, which replaces the reference by the output of an autoregressive teacher to reduce the ``modes'' (alternative translations for an input) in the training data. Besides, one thread of research explores the use of latent variables or word alignments to reduce the non-determinism in the translation process \cite{gu2017trainable,Ma_2019,Shu2020LatentVariableNN,song-etal-2021-alignart}, and another thread of research focuses on designing robust training objectives for NAT to mitigate the effect of multi-modality \cite{libovicky2018end,DBLP:conf/aaai/ShaoZFMZ20,Aligned,DBLP:conf/icml/DuTJ21}.

\begin{figure}[t]
  \begin{center}
    \includegraphics[width=1\columnwidth]{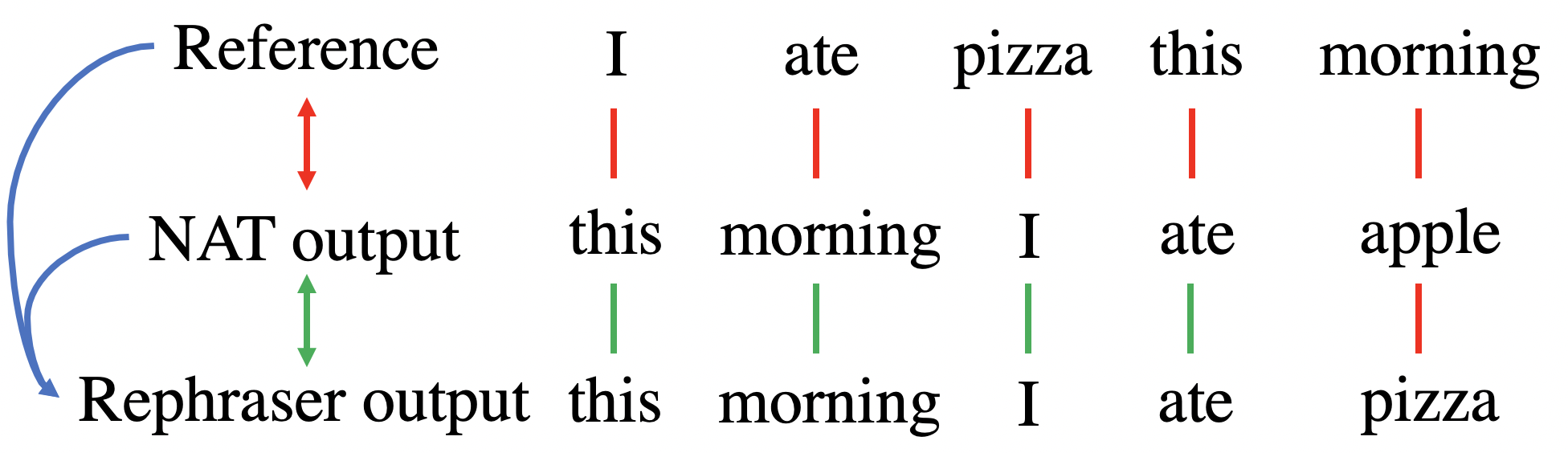}
    \caption{An example of the reference and the desired rephraser output. The reference is inappropriate for the training, and the rephraser output should keep the same semantics and correctly evaluate the model output.}
    \label{fig:example}
  \end{center}
\end{figure}
Our work handles the multi-modality problem in an unexplored way. Since the reference sentence may be inappropriate for the training, we introduce a rephraser to provide a better training target for NAT by rephrasing the reference sentence according to the NAT output. 
As Figure \ref{fig:example} shows, when the reference is ``I ate pizza this morning'' and the NAT output is ``this morning I ate apple'', although the only prediction error is `apple', predictions of all positions will be highly penalized by the inappropriate reference. If the reference can be rephrased to ``this morning I ate pizza'', then all predictions will be correctly evaluated, while the target-side semantics can still be preserved. 

How to obtain a good rephraser is the central focus of our approach. We use a shallow Transformer decoder as the rephraser. Since there is no direct supervision for the rephraser, we quantify our requirements for the rephraser as reward functions and optimize them with reinforcement learning. First, the rephraser output should fit well with the NAT output, otherwise an inappropriate reference will result in a large and inaccurate training loss. Therefore, the first reward is related to the training loss on the rephraser output. Second, the rephraser output should still maintain the original target-side semantics, so the second reward is the similarity between the reference sentence and rephraser output. Finally, we use an annealing strategy to find a balance between the two rewards. 

We evaluate our approach on major WMT benchmarks (WMT14 En$\leftrightarrow$De, WMT16 En$\leftrightarrow$Ro) and NAT baseline models (vanilla NAT, CMLM, CTC). Experimental results show that our approach consistently improves the translation quality of NAT. Specifically, our best variant achieves comparable performance to the autoregressive Transformer, while being 14.7 times more efficient in inference.
\section{Background}
\label{sec:2}
In this section, we briefly introduce different probability models for neural machine translation (NMT). We use $X$ to denote the source sentence and use $Y=\{y_1,...,y_T\}$ to denote the reference sentence.

\vspace{5pt}
\noindent{}\textbf{Autoregressive NMT}\ \ Autoregressive (AT) sequence models have achieved great success on machine translation, with different choices of architectures such as RNN \cite{bahdanau2014neural}, CNN \cite{gehring2017convolutional}, and Transformer \cite{vaswani2017attention}. Autoregressive models factorize the translation probability as follows and maximize it with the cross-entropy loss:
\begin{equation}
\mathcal{L}_{AT}(\theta) = -\sum_{t=1}^{T}\log(p(y_t|X,y_{<t},\theta)).
\end{equation}
During the inference, the preceding predicted tokens have to be fed into the decoder to generate the next token, which leads to the high translation latency of autoregressive NMT.

\vspace{5pt}
\noindent{}\textbf{Vanilla NAT}\ \ Non-autoregressive translation with the parallelizable Transformer architecture has been proposed to reduce the translation latency \cite{gu2017non}. NAT is able to generate all target tokens simultaneously since it breaks the sequential dependency in probability factorization. The vanilla NAT is also trained with the cross-entropy loss:
\begin{equation}
\label{eq:nat_mle}
\mathcal{L}_{NAT}(\theta) = -\sum_{t=1}^{T}\log(p(y_t|X,\theta)).
\end{equation}
The vanilla NAT is equipped with a length predictor that predicts the target length during the inference, and the translation of the given length is obtained by argmax decoding.

\vspace{5pt}
\noindent{}\textbf{CMLM}\ \ CMLM is an iterative decoding approach based on the conditional masked language model \cite{ghazvininejad2019maskpredict}. CMLM predicts the masked target tokens based on the source sentence and observed target tokens:
\begin{equation}
\mathcal{L}_{CMLM}(\theta) = -\log(P(Y_{mask}|X, Y_{obs},\theta)),
\end{equation}
where $Y_{mask}$ and $Y_{obs}$ represent the masked and observed target tokens respectively. During the training, $Y_{mask}$ is randomly selected among the target tokens. During the inference, the entire target sentence is masked in the first iteration, and then predictions with low confidence are masked and predicted again in the following iterations.

\vspace{5pt}
\noindent{}\textbf{CTC}\ \ Recently, CTC \cite{10.1145/1143844.1143891} is receiving increasing attention in NAT for its superior performance and the flexibility of variable length prediction \cite{libovicky2018end,saharia-etal-2020-non}. CTC-based NAT does not require a length predictor but generates an overlong sequence containing repetitions and blank tokens, which will be removed by a collapse function $\Gamma^{-1}$ in the post-processing to recover a normal sentence. CTC considers all sequences $A$ which the target sentence $Y$ can be recovered from, and marginalize the log-likelihood with dynamic programming:
\begin{equation}
 \mathcal{L}_{CTC}(\theta) = - \log \sum_{A \in \Gamma(Y)} p(A|X,\theta),
\end{equation}
where the probability $p(A|X,\theta)$ is modeled by a non-autoregressive Transformer.

\section{Approach}
In this section, we describe our approach in detail. First, we present the overview of our model, which incorporates the rephraser module into the encoder-decoder architecture. Then, we introduce the training methods for NAT with rephraser. Finally, we discuss the extensions of our approach on other major NAT baselines (i.e., CMLM and CTC). 

\begin{figure*}[t]
  \begin{center}
    \includegraphics[width=2.0\columnwidth]{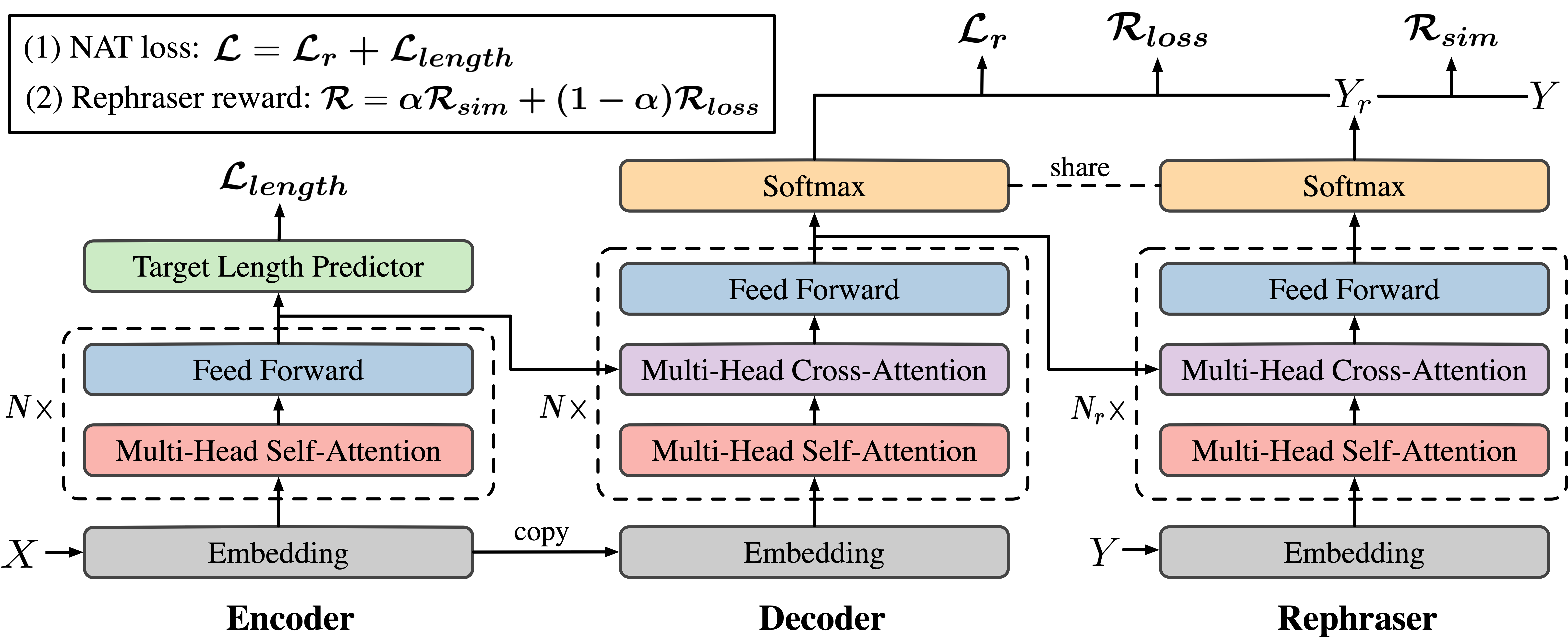}
    \caption{The architecture of the vanilla NAT with rephraser. The rephraser is stacked behind the decoder to provide a better training target for NAT. The cross-entropy loss, which is used to train the NAT model, is calculated based on the rephraser output rather than the reference sentence. The rephraser is trained to optimize two rewards, corresponding to the similarity with the reference sentence and the loss of NAT model respectively.}
    \label{fig:overview}
  \end{center}
\end{figure*}
\subsection{Model Overview}
\label{sec:31}
We present the overview of our model in Figure \ref{fig:overview}. In addition to the traditional encoder-decoder architecture, we introduce a rephraser module to rephrase the reference sentence according to the NAT output, which aims to provide a better training target for NAT. The rephraser only affect the training stage, so the inference cost remains the same. The total training cost of our approach is around 1.3 times compared to the NAT baseline.

As Figure \ref{fig:example} illustrates, we hope that the rephraser can rephrase the reference sentence into a form suitable for model training, which depends on both the reference sentence and NAT output. Therefore, we use the non-autoregressive Transformer decoder as the architecture of rephraser, which takes the reference $Y$ as the input and cross-attends to the output of NAT decoder. We use $Y_r=\{y_1^r,...,y_T^r\}$ to denote the rephraser output and use $P_r$ to denote the probability distribution of rephraser, which can be decomposed into the following form: 
\begin{equation}
 P_r(Y_r|X,Y,\theta) = \prod_{t=1}^{T}p_r(y_t^r|X,Y,\theta).
\end{equation}
Notice that here the rephraser does not change the target length, otherwise the cross-entropy loss will be inapplicable. We let the rephraser be a shallow model with the number of layers $N_r=2$ to reduce the additional training cost.

As Figure \ref{fig:overview} shows, we train NAT with the rephraser output $Y_r$ rather than the original reference $Y$. Specifically, we apply argmax decoding to obtain the rephraser output $Y_r$ and calculate the cross-entropy loss based on $Y_r$:
\begin{equation}
\mathcal{L}_{r}(\theta) = -\sum_{t=1}^{T}\log(p(y_t^r|X,\theta)).
\end{equation}

Since there is no direct supervision for the rephraser, we quantify our requirements for the rephraser as two reward functions and optimize them with reinforcement learning. The details will be discussed in the next section. 

\subsection{Training}
\label{sec:32}
In this section, we will describe in detail how to train the NAT model with a rephraser. We apply reinforcement learning to train the rephraser to produce better training targets for NAT. In order to make the learning of rephraser easier and avoid falling into local optima, we use a pre-training strategy to establish a good initial state for the rephraser. Since our method is sensitive to a hyperparameter that interpolates between two reward functions, we utilize a annealing strategy to find an optimal interpolation.

\subsubsection{Reinforcement Learning}
We apply reinforcement learning to train the rephraser since there is no direct supervision for it. The objective of the rephraser is to provide an appropriate training target for NAT, where the appropriateness can be quantified as two reward functions. 
As illustrated in Figure \ref{fig:example}, the rephraser output should fit well with the NAT output, otherwise the training loss for NAT will be dramatically overestimated. Therefore, we use a reward $\mathcal{R}_{loss}$ to encourage the reduction of training loss. Besides, the rephraser output should not change the target-side semantics, so we use a reward $\mathcal{R}_{sim}$ to measure the semantic similarity between the reference sentence and rephraser output.

Given the rephraser output $Y_r=\{y_1^r,...,y_T^r\}$, we define the reward $\mathcal{R}_{loss}$ as the negative training loss of NAT, which is the log-likelihood of $Y_r$:
\begin{equation}
\mathcal{R}_{loss}(Y_r) = \frac{\sum_{t=1}^{T}\log(p(y_t^r|X,\theta))}{T},
\end{equation}
where the length normalization keeps the scale of reward stable. We use $\mathcal{S}(Y_1,Y_2)$ to denote the similarity function that measures the semantic similarity between $Y_1$ and $Y_2$, and define the reward $\mathcal{R}_{sim}$ as:
\begin{equation}
\mathcal{R}_{sim}(Y_r) = \mathcal{S}(Y,Y_r).
\end{equation}
The similarity function can be chosen from the evaluation metrics in machine translation (e.g., BLEU \cite{papineni2002bleu}, METEOR \cite{banerjee-lavie-2005-meteor}, BERTScore \cite{zhang2019bertscore}), and we use BLEU in our experiments. We use a hyperparameter $\alpha \in [0,1]$ to interpolate between the two rewards for the rephraser:
\begin{equation}
\label{eq:rsim}
\mathcal{R}(Y_r) = \alpha \mathcal{R}_{sim}(Y_r)+(1-\alpha)\mathcal{R}_{loss}(Y_r).
\end{equation}

We apply the REINFORCE algorithm \cite{williams1992simple} to optimize the expected reward:
\begin{equation}
\begin{aligned}
\label{eq:rl}
\nabla_{\theta}&\mathcal{J}(\theta)=\nabla_{\theta}\sum_{Y_r}P_r(Y_r|X,Y,\theta)\mathcal{R}(Y_r)\\
&=\mathop{\mathbb{E}}_{Y_r \sim P_r}[\nabla_{\theta}\log P_r(Y_r|X,Y,\theta)\mathcal{R}(Y_r)].
\end{aligned}
\end{equation}
Specifically, we apply monte carlo sampling that samples a sentence $Y_r$ with its probability $P_r(Y_r|X,Y,\theta)$. In non-autoregressive models, it is equivalent to sampling a word $y_t$ in each position $t$, which is very efficient since the whole probability distribution is available. Then we calculate the reward $\mathcal{R}(Y_r)$ and update the rephraser with the estimated gradient $\nabla_{\theta}\log P_r(Y_r|X,Y,\theta)\mathcal{R}(Y_r)$. To reduce the variance of gradient estimation, we subtract a baseline reward from $\mathcal{R}(Y_r)$ \cite{10.5555/647235.720252}. In our setting, the baseline reward is obtained by sampling additional $K=2$ sentences and calculating their average reward.
\subsubsection{Pre-training}
If we directly train the rephraser with reinforcement learning, the rephraser will be trapped in a non-optimal state that it only outputs meaningless sentences composed of some most frequent words. This is a common problem of reinforcement learning since it is based on exploration and therefore sensitive to the initial state.

In order to make the learning of rephraser easier and avoid falling into local optima, we use a pre-training strategy to establish a good initial state for the rephraser. As the optimal rephraser output is usually a combination of the reference sentence and NAT output, we let the rephraser learn from both of them in the pre-training stage. Specifically, we use the average of two cross-entropy losses to train the rephraser, where the first is calculated based on the reference sentence and the second is calculated based on the argmax decoding result of NAT. As the rephraser is not ready for use, we train the NAT model with the original cross-entropy loss (Equation \ref{eq:nat_mle}) in the pre-training stage. 

\subsubsection{Annealing}
In reinforcement learning, we use a hyperparameter $\alpha$ to interpolate between the two rewards $\mathcal{R}_{loss}$ and $\mathcal{R}_{sim}$. The hyperparameter $\alpha$ is critical since it determines the behavior of the rephraser, but finding the optimal $\alpha$ with grid-search will dramatically increase the training cost. To solve this problem, we replace the expensive grid-search with an annealing strategy, which helps us efficiently find the optimal interpolation.

The behavior of the rephraser is determined by the value of $\alpha$, where the rephraser output is closer to the reference under a high $\alpha$ and closer to the NAT output under a low $\alpha$. It will be harmful to the NAT model if $\alpha$ is much lower than optimal, where the NAT model may become over-confident and easily collapse. In comparison, the impact of a high $\alpha$ is more moderate since the training target will be close to the reference sentence. Therefore, if we gradually decrease the value of $\alpha$ after the pre-training, it will act like a gradual fine-tuning process that the model gradually performs better until the optimal $\alpha$ is reached. With this strategy, we only need to select checkpoints on the validation set to find the optimal interpolation.

Specifically, let the number of fine-tuning steps be $T$ and the current step be $t$. We use a linear annealing schedule to gradually decrease $\alpha$ from $\alpha_{max}$ to $\alpha_{min}$ in the fine-tuning:
\begin{equation}
\label{eq:alpha}
\alpha = \frac{t}{T}\alpha_{min}+(1-\frac{t}{T})\alpha_{max}.
\end{equation}
Though we replace $\alpha$ with two hyperparameters $\alpha_{max}$ and $\alpha_{min}$, the model performance is not very sensitive to them. Therefore, we do not need to tune them carefully and can use the same set of hyperparameters across different datasets.
\subsection{Extensions}
\label{sec:33}
Currently, we have discussed how to incorporate the rephraser module into the vanilla NAT. Actually, our approach is a general framework that does not assume a specific NAT model. In this section, we extend our approach to CMLM and CTC, which are major NAT baselines introduced in section \ref{sec:2}.

\vspace{5pt}
\noindent{}\textbf{CMLM} The main difference between CMLM and vanilla NAT is that CMLM has observed some target tokens and only predicts the masked tokens in the training. Therefore, we let the observed tokens remain unchanged and only rephrase the reference of the masked tokens.

\vspace{5pt}
\noindent{}\textbf{CTC} The major feature of CTC-based NAT is the flexibility of variable length prediction, which allows the rephraser to change the target length. Therefore, we use a CTC-based rephraser structure, which takes the decoder output as input and cross-attends to the embedding of the reference sentence $Y$. Similarly, we remove the repetitions and blank tokens to obtain the rephraser output $Y_r$ and calculate the probability $P_r(Y_r)$ with dynamic programming.

\section{Experiments}
\subsection{Experimental Setup}
\begin{table*}[th!]
\centering
\small
\begin{tabular}{clcccccc}
\toprule
 \multirow{2}{*}{\textbf{Models}} && \multirow{2}{*}{\textbf{Iter}} & \multirow{2}{*}{\textbf{Speed}} &
 \multicolumn{2}{c}{\textbf{WMT14}} & \multicolumn{2}{c}{\textbf{WMT16}} \\
  &&&&\textbf{EN-DE} & \textbf{DE-EN} & \textbf{EN-RO} & \textbf{RO-EN} \\
\midrule
\multirow{3}{*}{AT}
& Transformer (teacher) & $N$& 1.0$\times$&27.54&\bf{31.57}&\bf{34.26}&\bf{33.87} \\
& Transformer (12-1) &$N$& 2.6$\times$&  26.09&30.30&32.76&32.39  \\
& \ \ \ + KD &$N$& 2.7$\times$&  \bf{27.61}&31.48&33.43&33.50  \\
\cmidrule[0.6pt](lr){1-8}
\multirow{6}{*}{Vanilla NAT}
&NAT-FT \cite{gu2017non} &$1$&15.6$\times$&17.69 & 21.47 & 27.29 &  29.06 \\
&Bag-of-ngrams \cite{DBLP:conf/aaai/ShaoZFMZ20} &1&10.7$\times$&20.90 &24.61 &28.31 &29.29\\
&EM \cite{pmlr-v119-sun20c} & 1 & 16.4$\times$ & 24.54 & 27.93 & -- & --\\
&GLAT \cite{qian-etal-2021-glancing} &$1$&15.3$\times$&25.21&\bf{29.84}&31.19&\bf{32.04}\\
&Vanilla NAT (ours)&$1$&15.6$\times$&20.42 &24.88& 29.21& 29.37\\
&Vanilla NAT w/ rephraser &$1$&15.6$\times$& \bf{25.33}& 29.57& \bf{31.63}& 31.72\\
\cmidrule[0.6pt](lr){1-8}
\multirow{6}{*}{CMLM}
& CMLM \cite{ghazvininejad2019maskpredict} & $1$&--&18.05 & 21.83 & 27.32 & 28.20 \\
& CMLM + DSLP \cite{huang2022non} &$1$&15.0$\times$& 21.76 & 25.30 & 30.29 & 30.89 \\
& CMLM + AXE \cite{Aligned} &$1$&--& 23.53 & 27.90 & 30.75 & 31.54 \\
& CMLM + OAXE \cite{DBLP:conf/icml/DuTJ21} &$1$&--&26.10 & 30.20 & 32.40 & \bf{33.30}\\
& CMLM (ours) &$1$&15.0$\times$& 18.21& 22.86&28.15 &28.97 \\
& CMLM w/ rephraser &$1$&15.0$\times$&\bf{26.65} &\bf{30.70}& \bf{32.72}&33.03 \\
\cmidrule[0.6pt](lr){1-8}
\multirow{9}{*}{CTC}
& CTC \cite{libovicky2018end}  &$1$&--& 16.56 & 18.64 & 19.54 & 24.67  \\
& Imputer \cite{saharia-etal-2020-non}  &$1$&--& 25.80 & 28.40 & 32.30 & 31.70 \\
& REDER \cite{zheng2021duplex}  &$1$&15.5$\times$&26.70 & 30.68 & 33.10 & 33.23\\
& Fully-NAT \cite{gu2020fully}  &$1$&16.8$\times$& 27.20 & 31.39 & 33.71 & 34.16\\
& NMLA \cite{nmla}& 1 & 14.7$\times$& 27.57 & 31.28 & 33.86 & 33.94\\
& DDRS \cite{shao-etal-2022-one}& 1 & 14.7$\times$&27.60&31.48&\bf{34.60}&\bf{34.65}\\
& CTC (ours) &$1$&14.7$\times$&26.34 & 29.58 & 33.45 & 33.32\\
& CTC w/ rephraser &$1$&14.7$\times$& 27.32& 30.97&33.80 &33.84 \\
& NMLA w/ rephraser &$1$&14.7$\times$& \bf{27.81}& \bf{31.53}&34.08 &34.03 \\
\bottomrule
\end{tabular}
\caption{Performance comparison between our models and existing methods. AT means autoregressive. `12-1' means the Transformer with 12 encoder layers and 1 decoder layer. `Iter' means the number of decoding iterations.}
\label{tab:main-rst}
\end{table*}

\begin{table*}[h]
\centering
\small
\begin{tabular}{lccccc}
\toprule
 \multirow{2}{*}{\textbf{Models}} & \multirow{2}{*}{\textbf{Iter}} &
 \multicolumn{2}{c}{\textbf{WMT14}} & \multicolumn{2}{c}{\textbf{WMT16}} \\
  &&\textbf{EN-DE} & \textbf{DE-EN} & \textbf{EN-RO} & \textbf{RO-EN} \\
\midrule
CMLM + AXE \cite{Aligned} &$1$& 20.40 & 24.90 & 30.47 & 31.42 \\
CMLM + OAXE \cite{DBLP:conf/icml/DuTJ21} &$1$&22.40 & 26.80 & -- & --\\
CMLM (ours) &$1$& 10.82& 14.64&23.51 &24.25 \\
CMLM w/ rephraser &$1$&\bf{23.12} &\bf{27.44}& \bf{32.30}&\bf{32.07} \\
\bottomrule
\end{tabular}
\caption{The performance of CMLM with rephraser and other existing methods on raw data. `--' means not reported.}
\label{tab:raw}
\end{table*}
\noindent{}\textbf{Data} We conducted experiments on major benchmarking datasets in previous NAT studies: WMT14 English$\leftrightarrow$German (En$\leftrightarrow$De, 4.5M sentence pairs) and WMT16 English$\leftrightarrow$Romanian (En$\leftrightarrow$Ro, 0.6M sentence pairs). For WMT14 En$\leftrightarrow$De, the validation set is \textit{newstest2013} and the test set is \textit{newstest2014}. For WMT16 En$\leftrightarrow$Ro, the validation set is \textit{newsdev-2016} and the test set is \textit{newstest-2016}. We learn a joint BPE model \cite{sennrich2015neural} with 32K merge operations to process the data and share the vocabulary for source and target languages. We use BLEU \cite{papineni2002bleu} to evaluate the translation quality. 

\vspace{5pt}
\noindent{}\textbf{Knowledge Distillation} We follow previous works on NAT to apply sequence-level knowledge distillation \cite{kim-rush-2016-sequence} to reduce the complexity of training data. We employ the base version of autoregressive Transformer \cite{vaswani2017attention} as the teacher model and train NAT models on the translations generated by teachers.

\vspace{5pt}
\noindent{}\textbf{NAT Baselines} We conduct experiments on three major NAT baselines: vanilla NAT \cite{gu2017non}, CMLM \cite{ghazvininejad2019maskpredict}, and CTC \cite{libovicky2018end}. We adopt Transformer-base as the model architecture. For vanilla NAT and CTC, we apply the uniform copy \cite{gu2017non} to construct decoder inputs. For CTC, we also evaluate the rephraser performance when the non-monotonic latent alignments (NMLA) objective \cite{nmla} is applied for fine-tuning.

\vspace{5pt}
\noindent{}\textbf{Implementation Details} We set $\alpha_{max}$ to 0.75, $\alpha_{min}$ to 0.5, the sampling times $K$ to 2, and the depth of rephraser $N_r$ to 2 across all datasets. For CTC, the length for decoder inputs is $3\times$ as long as the source length. All models are optimized with Adam \cite{DBLP:journals/corr/KingmaB14} with $\beta=(0.9,0.98)$ and $\epsilon=10^{-8}$. For vanilla NAT and CTC, each batch contains approximately 64K source words. For CMLM, to keep consistency with previous works \cite{ghazvininejad2019maskpredict,Aligned,DBLP:conf/icml/DuTJ21}, we use the batch size 128K and use 5 length candidates for inference. All models are pre-trained for 300K steps and fine-tuned for 30K steps. During the fine-tuning, we measure validation BLEU for every 500 steps and average the 5 best checkpoints to obtain the final model. We use the GeForce RTX $3090$ GPU to train models and measure the translation latency. We implement our models based on the open-source framework of fairseq \cite{ott2019fairseq}.
\subsection{Main Results}
\begin{figure}[t]
    \centering
    \subfloat[WMT14 En-De]{
    \includegraphics[height=0.188\textwidth]{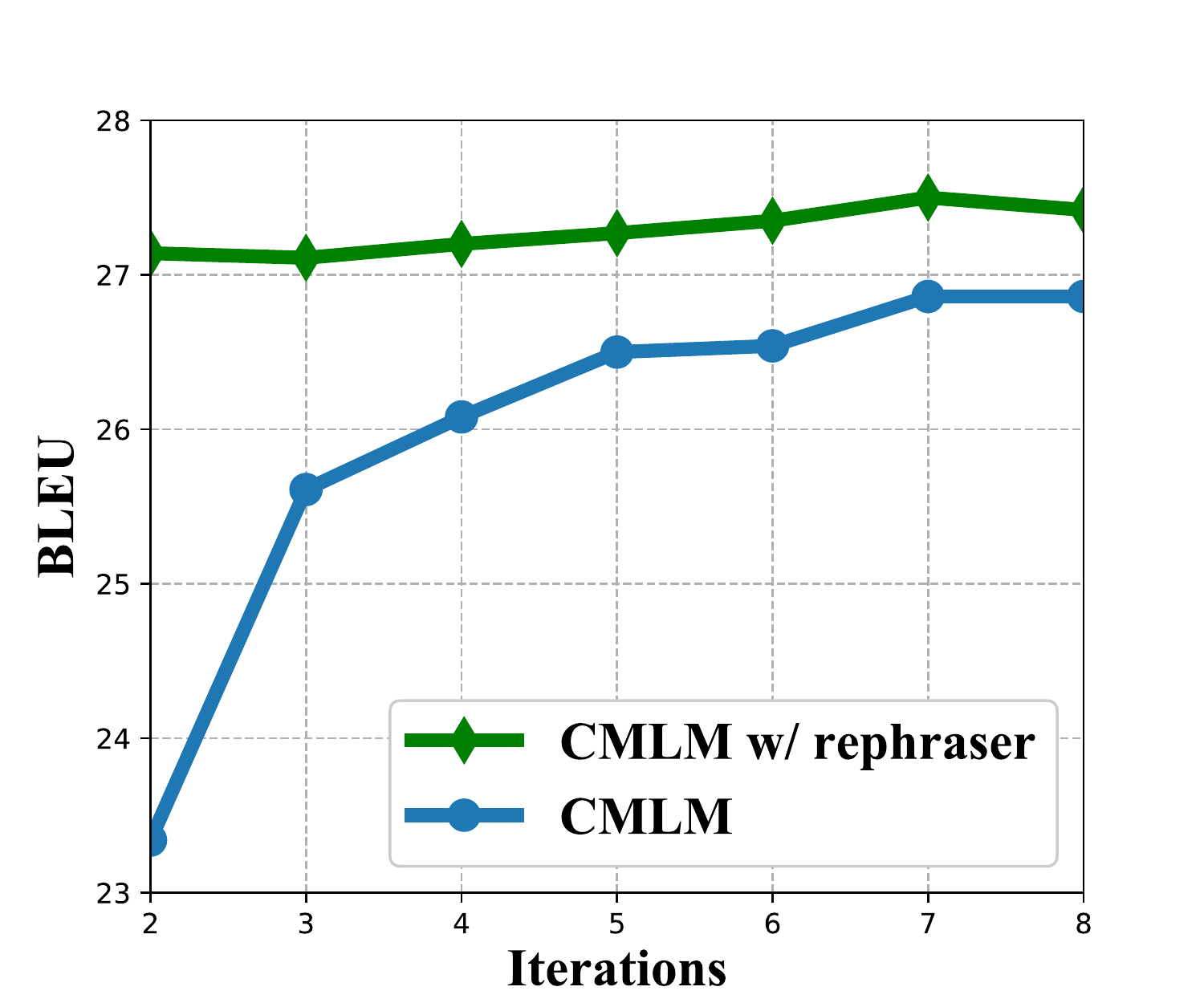} \label{fig:ende}
    } \hfill
    \subfloat[WMT14 De-En]{
    \includegraphics[height=0.188\textwidth]{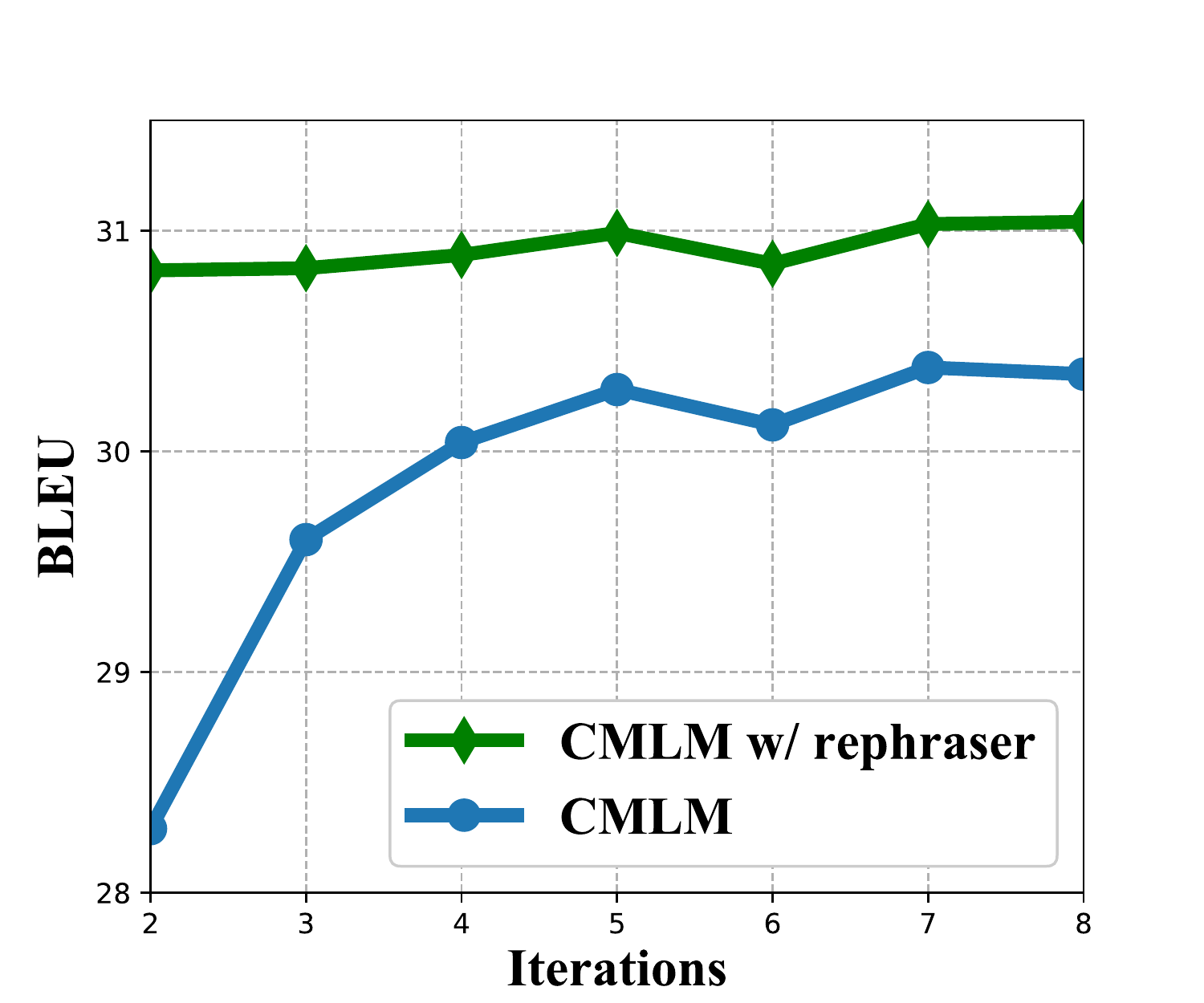} \label{fig:deen}
    }
    \caption{Performance comparison between CMLM and CMLM with rephraser under different iterations.}
    \label{fig:iter}
\end{figure}

We report the performance of our models and existing one-iteration NAT approaches in Table \ref{tab:main-rst}. By rephrasing the reference sentence, our approach improves all baseline models by a large margin and achieves comparable performance to the state-of-the-art method on each baseline model. Notably, our approach achieves an average improvement of 6.2 BLEU on CMLM and 3.6 BLEU on vanilla NAT. The rephraser also works well on the strong baseline of CTC and NMLA and achieves the comparable performance to autoregressive Transformer while being much faster in inference.

We also evaluate the effectiveness of our approach on iterative decoding. In Figure \ref{fig:iter}, we report the performance of CMLM models under different iterations. On both WMT14 En-De and WMT14 De-En test sets, CMLM with rephraser consistently outperforms the CMLM baseline, showing that the rephraser is also helpful for iterative NAT. Notably, our model with only 2 decoding iterations can outperform the CMLM baseline of 8 iterations.

The above models are trained with the help of sequence-level knowledge distillation, which replaces the reference with the output of a teacher model. As our rephraser is also capable of modifying the reference, we conduct experiments on raw data to see whether our approach can be used without knowledge distillation. In Table \ref{tab:raw}, we report the performance of rephraser and other existing methods on the CMLM baseline. CMLM with rephraser significantly improves the baseline by more than 10 BLEU on average and also outperforms other existing methods. 
Compared with the knowledge distillation variant, the performance degradation is about 3.5 BLEU on WMT14 En$\leftrightarrow$De and less than 1 BLEU on WMT16 En$\leftrightarrow$Ro. Though the rephraser cannot replace knowledge distillation yet, the relatively small performance degradation shows the potential in this direction.

\subsection{Ablation Study}
In this section, we conduct ablation studies on WMT14 En-De validation set to justify the settings in the main experiment, including the model size, rephraser architecture, reward function, and annealing strategy.

\vspace{5pt}
\noindent{}\textbf{Model Size \& Training Steps} Our approach requires 2 extra decoder layers for the rephraser module and 30K extra training steps for the rephraser fine-tuning. Considering that these factors may also strengthen the NAT baseline, we conduct experiments on CTC to study their effects. As Table \ref{tab:addone} shows, the extra decoder layers and training steps only have marginal effects on the model performance. In comparison, CTC with rephraser outperforms these variants, and the rephraser does not affect the decoding speed.

\begin{table}[t]
    \small
    \centering
    \begin{tabular}{cc|ccc}
        \toprule
        \multicolumn{2}{c|}{CTC}& \multirow{2}{*}{{Params}} & \multirow{2}{*}{{Speed}} & \multirow{2}{*}{BLEU}\\
        +2 layers&+30K steps& &  & \\
        \midrule
        &&62.10M&14.7$\times$&24.77\\
        \checkmark&&70.51M&12.9$\times$&24.89\\
        &\checkmark&62.10M&14.7$\times$&24.80\\
        \checkmark&\checkmark&70.51M&12.9$\times$&24.87\\
        \midrule
        \multicolumn{2}{c|}{CTC w/ rephraser}&70.51M&14.7$\times$&25.54\\
    \bottomrule
    \end{tabular}
        \caption{The effect of 2 extra decoder layers and 30K extra training steps on the CTC model. `Params' means the number of parameters. `Speed' means the speedup to autoregressive Transformer under batch size 1.}
    \label{tab:addone}
\end{table}

\vspace{5pt}
\noindent{}\textbf{Rephraser Architecture} As the rephraser is the core of our approach, its architecture will have a certain impact on model performance. Our rephraser is a Transformer decoder that is fed the reference $Y$ and cross-attends to the output of NAT decoder. There are other choices of the rephraser architecture that has the NAT output and reference $Y$ as inputs. For example, we can swap the two inputs of the rephraser and we call this architecture `swap', which is fed the NAT output and cross-attends to the reference $Y$. We can also compress the two inputs into one with a feed-forward layer and then feed it into a Transformer encoder, and we call this architecture `enc'. Besides, we can change the depth of the rephraser, and we try two choices $N_r=1$ and $N_r=4$. We try these variants on vanilla NAT and report the results in Table \ref{tab:archi}.
\begin{table}[t]
\small
\centering
\begin{tabular}{lccccc}
\toprule
{\textbf{Models}} &base& swap & enc& $N_r=1$&$N_r=4$\\
\midrule
BLEU&24.06&24.01&23.61&23.53&24.09\\
\bottomrule
\end{tabular}
\caption{The performance of vanilla NAT with different rephraser architectures.}
\label{tab:archi}
\end{table}

We can see that the Transformer decoder architecture is better than the encoder, and swapping the two inputs of the decoder does not have an obvious effect on the model performance. Regarding the depth of rephraser, increasing $N_r$ to 4 does not make a big difference, but reducing $N_r$ to 1 will certainly degrade the model performance.

\vspace{5pt}
\noindent{}\textbf{Reward Function} In the default setting, we use BLEU to measure the similarity between the reference and rephraser output. 
As our approach is capable of incorporating different evaluation metrics as reward functions, we investigate the effect of two other reward functions METEOR \cite{banerjee-lavie-2005-meteor} and BERTScore \cite{zhang2019bertscore} on vanilla NAT. As shown in Table \ref{tab:reward}, the BLEU reward performs better than METEOR and BERTScore under all these three evaluation metrics, which is a little surprising since BERTScore is a better evaluation metric based on pre-trained models. We speculate that the strength of BLEU lies in its robustness, which is solely based on the statistical n-gram matching. 

\begin{table}[t]
\small
\centering
\begin{tabular}{lccc}
\toprule
{\textbf{Metrics}} &BLEU& METEOR & BERTScore\\
\midrule
BLEU&24.06&52.88&84.24\\
METEOR&23.71&52.65&84.11\\
BERTScore&23.07&52.02&83.80\\
\bottomrule
\end{tabular}
\caption{The performance of vanilla NAT with different reward functions for the rephraser.}
\label{tab:reward}
\end{table}

\vspace{5pt}
\noindent{}\textbf{Annealing} We replace the static $\alpha$ with a annealing strategy from $\alpha_{max}$ to $\alpha_{min}$, which is less sensitive to hyperparameters. To justify this claim, we deviate the hyperparameters $\{\alpha,\alpha_{max},\alpha_{min}\}$ from the optimal value by $\Delta$ and report how the model performance is affected in Table \ref{tab:curr}. Though the model is sensitive to the static $\alpha$, the performance of annealing is almost unaffected by a small deviation of hyperparameters, which well supports our claim.
\begin{table}[t]
\small
\centering
\begin{tabular}{lccc}
\toprule
$\Delta$ &0& 0.05 & -0.05\\
\midrule
Anneal&24.06&24.03&23.88\\
Static&23.97&23.55&23.63\\
\bottomrule
\end{tabular}
\caption{The performance of vanilla NAT with deviation $\Delta$ from the optimal setting. `Anneal' represents annealing and `Static' represents the static $\alpha$.}
\label{tab:curr}
\end{table}


\subsection{Analysis}
In this section, we present a qualitative analysis on generated outputs of the WMT14 En-De test set to better understand where the improvements come from and whether the rephraser can alleviate the multi-modality problem.

\vspace{5pt}
\noindent{}\textbf{Prediction Confidence} Due to the multi-modality problem, NAT may consider many possible translations at the same time, which makes NAT less confident in generating outputs. Therefore, we compare the prediction confidence between baseline models and our models to investigate whether the rephraser can alleviate the multi-modality problem. We use the information entropy $H(X)\!=\!-\!\sum_{x}p(x)\log p(x)$ to measure the prediction confidence, where lower entropy indicates higher confidence. Table \ref{tab:entropy} shows that NAT models with rephraser have much lower entropy than baseline models, illustrating the effectiveness of rephraser in alleviating the multi-modality problem.
\begin{table}[t]
\small
\centering
\begin{tabular}{lccc}
\toprule
{\textbf{Models}} & {\textbf{Vanilla NAT}} & {\textbf{CMLM}} & {\textbf{CTC}}\\
\midrule
w/o rephraser&2.07&2.64&0.26\\
w rephraser&1.35&1.52&0.06\\
\bottomrule
\end{tabular}
\caption{The average entropy of NAT models. Lower entropy indicates higher prediction confidence.}
\label{tab:entropy}
\end{table}

\vspace{5pt}
\noindent{}\textbf{Token Repetitions} With low prediction confidence, NAT may generate a mixture of many possible translations, which makes the translation inconsistent and typically contains many repetitive tokens. Therefore, we report the percentage of token repetitions in Table \ref{tab:rep}, which shows that our approach can significantly reduce token repetitions. Notice that we do not report the results of CTC since it naturally contains no token repetitions.

\begin{table}[t]
\small
\centering
\begin{tabular}{lcc}
\toprule
{\textbf{Models}} & {\textbf{Vanilla NAT}} & {\textbf{CMLM}}\\
\midrule
w/o rephraser&10.2\%&15.4\%\\
w rephraser&2.6\%&2.3\%\\
\bottomrule
\end{tabular}
\caption{The percentage of token repetitions in the generated outputs of Vanilla NAT and CMLM.}
\label{tab:rep}
\end{table}

\subsection{Case Study}
We present a case of the rephraser output in Table \ref{tab:case} to better understand the working mechanism of rephraser. Before rephrasing, the reference is not aligned with NAT output since there exists the multi-modality problem (``but our loyalty is clear.'' and ``but it is not about our loyalty.''). After rephrasing, the reference basically keeps the original semantics and is well aligned with NAT output. 

\begin{table}[h]
\small
\centering
\begin{tabular}{c|l}
\toprule
\multirow{2}{*}{Reference} & Our identity might sometimes be fu@@ \\ &zz@@ y , but our loy@@ alty is clear .\\
\hline
\multirow{2}{*}{NAT} & Our identity may sometimes be urred ,\\& but it is not about our loy@@ alty .\\
\hline
\multirow{2}{*}{Rephraser} & Our identity might sometimes be vague ,\\& but that is not about our loy@@ alty .\\
\bottomrule
\end{tabular}
\caption{A case of the rephraser output from the validation set of WMT14 De-En.}
\label{tab:case}
\end{table}
\section{Related Work}
\citet{gu2017non} first proposed non-autoregressive Transformer to reduce the latency of machine translation by generating outputs in parallel. However, there is a large performance gap between the AT and the earliest NAT model, which is mainly attributed to the multi-modality problem.

A number of efforts have explored ways to help NAT handle the multi-modality problem. One thread of research reduces the modality in the target space by leaking part of the target information in the training. \citet{kaiser2018fast,Ma_2019,Shu2020LatentVariableNN,bao-etal-2021-non} augmented NAT with latent variables based on vector quantization or variational inference. Besides, \citet{gu2017non,ran2019guiding,bao2019nonautoregressive,song-etal-2021-alignart} explored the use of word alignments to help the generation of NAT, and \citet{akoury-etal-2019-syntactically,liu-etal-2021-enriching} enriched NAT with syntactic structures.

Another thread of research focuses on designing robust training objectives for NAT to mitigate the effect of multi-modality. \citet{wang2019non} introduced regularization terms to reduce errors of repeated and incomplete translations. \cite{tu-etal-2020-engine} viewed NAT as an inference network trained to minimize the autoregressive teacher energy. \citet{shao-etal-2019-retrieving,DBLP:conf/aaai/ShaoZFMZ20,DBLP:journals/corr/abs-2106-08122} introduced sequence-level training objectives for NAT. \citet{Aligned,DBLP:conf/icml/DuTJ21} improved the cross-entropy loss with better alignments. Recently, the CTC loss \cite{10.1145/1143844.1143891} is receiving increasing attention in NAT \cite{libovicky2018end,saharia-etal-2020-non,gu2020fully,nmla}, which is further enhanced with a directed acyclic graph to explicitly model the probability of transistion paths \cite{huang2022directed,shao2022viterbi}.

Our work is closest to diverse distillation \cite{shao-etal-2022-one}, which provides multiple references in the dataset and dynamically select one reference to train the model. Diverse distillation has a high distillation cost, and the provided references cannot ensure that all possible outputs are covered. Our work overcomes these limitations by directly rephrasing the reference, which has a lower cost and can provide a more flexible reference.

We train the rephraser with the reinforcement learning technique \cite{williams1992simple}, which is widely used in neural machine translation to optimize sequence-level objectives like BLEU \cite{ranzato2015sequence,bahdanau2016actor,wu-etal-2018-study,DBLP:journals/corr/abs-2106-08122}. There are also works on paraphrasing the reference to provide a better training target for NMT \cite{sekizawa-etal-2017-improving,zhou-etal-2019-paraphrases,freitag-etal-2020-human}, where the paraphrasing is conducted by human or a paraphrase dictionary.

\section{Conclusion}
Since the reference may be inappropriate for the training of NAT, we propose rephrasing the reference to provide a better training target for NAT. We apply reinforcement learning to obtain a good rephraser and train NAT based on the rephraser output. Experiments on major benchmarks and NAT baselines demonstrate the effectiveness of our method.
\section{Acknowledgement}
We thank the anonymous reviewers for their thorough review and valuable feedback.
\bibliography{custom}
\end{document}